\documentclass[11pt]{article}

\usepackage{fullpage}
\usepackage[ruled, linesnumbered, vlined, commentsnumbered]{algorithm2e}
\usepackage{graphicx,subfig} % figure related
\usepackage{amsfonts,amssymb,amsmath,amsthm,amsopn}	% math related
\usepackage{hhline,booktabs,colortbl,multirow,tabularx,threeparttable} % table related
\usepackage[listings,skins,breakable]{tcolorbox}

\usepackage{enumerate}
\usepackage{authblk}
\usepackage{footnote}
\usepackage{hyperref}
\usepackage{prettyref}
\usepackage{cite}
\usepackage{setspace}
\usepackage{color}
\usepackage{xcolor}  % Required for custom colors
\usepackage{geometry}

\usepackage{tikz}
\usepackage{pgfplots}
\usetikzlibrary{positioning,shapes,shadows,arrows,calc}
\tikzstyle{component}=[rectangle, draw=black, rounded corners, fill=blue!40, drop shadow, text centered, anchor=north, text=white, minimum height=1cm]
\tikzstyle{arrow}=[->, thick]

\pgfplotsset{compat=1.12}
\usetikzlibrary{intersections}
\usetikzlibrary{pgfplots.statistics}
\usepgfplotslibrary{fillbetween}

\def\hlinew#1{%
  \noalign{\ifnum0=`}\fi\hrule \@height #1 \futurelet
   \reserved@a\@xhline}
\makeatother

\definecolor{myblue}{RGB}{34,31,217}
\definecolor{mycyan}{gray}{.7}
\definecolor{Gray}{gray}{0.9}

\newrefformat{fig}{Fig.~\ref{#1}}
\newrefformat{tab}{Table~\ref{#1}}
\newrefformat{sec}{Section~\ref{#1}}
\newrefformat{app}{Appendix~\ref{#1}}
\newrefformat{alg}{Algorithm~\ref{#1}}
\newrefformat{property}{Property~\ref{#1}}
\newrefformat{theorem}{Theorem~\ref{#1}}
\newrefformat{corollary}{Corollary~\ref{#1}}
\newrefformat{proposition}{Proposition~\ref{#1}}
\newrefformat{def}{Definition~\ref{#1}}
\newrefformat{eq}{equation~(\ref{#1})}

\title{\vspace{-1ex}\LARGE\textbf{Multi-Objective Reinforcement Learning based Multi-Microgrid System Optimisation Problem}\footnote{This manuscript is accepted for publication in EMO 2021. The copyright is transferred to Springer.}}

\author[1]{\normalsize Jiangjiao Xu}
\author[1]{\normalsize Ke Li}
\author[2]{\normalsize Mohammad Abusara}
\affil[1]{\normalsize Department of Computer Science, University of Exeter, EX4 4QF, Exeter, UK}
\affil[2]{\normalsize Department of Engineering, University of Exeter, EX4 4QF, Exeter, UK}
\affil[$\ast$]{\normalsize Email: \texttt{k.li@exeter.ac.uk}}
\date{}

\begin{document}
\maketitle

\vspace{-3ex}
{\normalsize\textbf{Abstract: }Microgrids with energy storage systems and distributed renewable energy sources play a crucial role in reducing the consumption from traditional power sources and the emission of $CO_2$. Connecting multi microgrid to a distribution power grid can facilitate a more robust and reliable operation to increase the security and privacy of the system. The proposed model consists of three layers, smart grid layer, independent system operator (ISO) layer and power grid layer. Each layer aims to maximise its benefit. To achieve these objectives, an intelligent multi-microgrid energy management method is proposed based on the multi-objective reinforcement learning (MORL) techniques, leading to a Pareto optimal set. A non-dominated solution is selected to implement a fair design in order not to favour any particular participant. The simulation results demonstrate the performance of the MORL and verify the viability of the proposed approach.
}

{\normalsize\textbf{Keywords: } }Multi-microgrid,  multi-objective reinforcement learning, independent system operator, market operator, Pareto Front.

\section{Introduction}
% no \IEEEPARstart
Over the past decade, the leading domestic electricity price paid by households is fixed for their energy consumption, predominantly gas and electricity. At the same time, non-hydro renewable energy sources (RES), such as wind, solar, tidal and geothermal power, continues to penetrate the power generation market in meaningful ways. The percentage of these sources has risen from 0.37 $\%$ in 1990 to 5.93 $\%$ in 2016 \cite{1}. It is well understood that wholesale price variability is an essential characteristic of deregulation in the electricity market. Energy consumers who are sensitive to energy prices may vary their electricity consumption based on the dynamic price signals \cite{2}. It means that dynamic electricity prices can reduce the demand for peak load period, while the use of renewable energy and energy storage systems can significantly reduce the use of fossil fuels, thereby reducing power generation costs and carbon dioxide emissions.

A wide range of demand side management studies in dynamic pricing schemes have been conducted. Reference \cite{3} presented a demand response approach based on the dynamic energy pricing, which accomplishes the optimal load control of the devices by establishing a virtual power transaction process.  A decision model for smart grid considering the demand response and market energy pricing is proposed to interact between the market retail price and energy consumers \cite{4}. In \cite{5}, to improve the system operation and optimise the power flow, a coordinated operating strategy for the gas and electricity integrated energy system is proposed in a multi-energy system. However, the above studies in demand side management only optimise the energy prices from an operational perspective and do not acknowledge the influence of price variations in the energy market and customer demand on the planning level. Moreover, the majority of existing articles study single utility objective optimisation problem solely, e.g., reduce the overall cost \cite{6}, maximise costumers' utility \cite{7} and modelling demand figure \cite{8}. When a multi-microgrid system is designed, there will be a coupling interaction between microgrids, independent system operator (ISO) and the power grid. There are often some conflicts between these participants during planning. The impact of dynamic pricing in a multi-microgrid system for a multi-objective problem has not been investigated comprehensively.

To investigate a comprehensive model and balance all participants, we consider designing a multi-microgrid system, including three microgrids, an independent system operator (ISO) and a main power grid \cite{9}. Microgrids are connected to each other, but renewable energy generation cannot be dispatched between each other. A dynamic pricing scheme will be implemented in this multi-microgrid system to balance the system operation for all participants. In this case, a multi-objective optimal approach needs to be proposed to balance all objectives without favour any single participant. 

Multi-Objective Reinforcement learning is an outstanding algorithm to address multi-objective problems for complex strategic interactions. In \cite{10}, a reinforcement learning environment is typically formalised by employing a Markov decision process (MDP). A Q-learning algorithm was introduced to approximate optimal $Q$ values in \cite{11} iteratively. In a multi-objective optimisation problem, the objective space will contain two or more dimensions, and regular MDPs will be generalised to multi-objective MDPs. Many approaches of MORL rely on a single-policy algorithm to learn Pareto optimal solutions \cite{13}. The most straightforward idea is to convert the multi-objective problem into a standard single-objective problem by utilising a scalarisation function \cite{LiDZK15,ChenLY18,LiCFY19}.

However, this transformation may not be suitable to solve a non-linear problem that lies in non-convex regions of the Pareto front. In this paper, we develop a $L_p$ metrics based MORL algorithm to create a fair multi-microgrid system design based on the Approximate Pareto Front (APF), which can achieve high-quality solutions to non-linear multi-objective functions. To the best of our knowledge, the proposed MORL is the first time to be used in a multi-objective optimisation smart grid scenario. Then the Pareto Front will be applied to investigate the connection and influence between various objective functions, which can support to produce a fair result for all participants. Our main contributions are:
%
%\begin{enumerate}[]
%
%\item 

It combines real-time various energy prices with planning scenarios in practice and considers the impact of real-time changing electricity prices and renewable energy on the design of multi-microgrid systems by implementing a method of a MORL algorithm. The main grid provider sales revenue, the university office utilises the renewable energy to operate the energy storages and the users save energy consumption cost. Three conflict objectives are included in the multi-microgrid planning scenario.

%\item 
A multi-objective formulation is developed to a future multi-microgrid framework. To solve a multi-objective problem (MOP), we develop a MORL algorithm considering the dynamic electricity prices and the operation of energy storage, e.g., charging/discharging/idle, which can generate an APF to provide a fair and effectiveness operating planning for a multi-microgrid network.

%\end{enumerate}

The rest of this report is organised as follows. Section II describes the main structure of the multi-microgrid network and discusses the mathematical system models of three participants. The proposed MORL problem formulation is presented in Section III, and Section IV describes the methodology of MORL algorithms in detail. In Section V, the numerical simulation results based on the Pareto are presented. Finally, section VI is the conclusion and future work.

\section{Multi-Microgrid Description}

A multi-microgrid system based on the Penryn Campus, University of Exeter is shown in Fig. 1. It consists of RESs, Microgrids, independent system operator and Power Grid. Information and communication technology (ICT) systems are implemented to exchange the information among microgrids, i.e., price, power demand and generation. A high-level multi-microgrid optimisation system is considered to be designed in this paper. Detailed mathematical models of the power grid, ISO and microgrid will be presented as follows. Let $\mathcal{N}={1,2,...,N}$ and $\mathcal{N}_s={1,2,...,N_s}$ denote the set of the microgrids and the set of the microgrids with energy storages, respectively, where $N_s \leq N$.

\subsection{Microgrid Model}

The mathematical models of a microgrid model will show the power balance among energy storages, microgrids and the power grid. For microgrid $n$ without energy storage, it has

\vspace{-0.6mm}
\begin{equation}
\begin{array}{rcl}
p_{d_n}(t)=p_{g_n}(t)+p_{r_n}(t).
\end{array}
\end{equation}

\noindent where $p_{g_n}(t)$ is the power transmission between microgrid $n$ and the power grid at time $t$. If $p_{g_n}(t)$ is positive, the power grid will transmit the power to the microgrid $n$, otherwise, the microgrid $n$ will send the power back to the grid. $p_{r_n}(t)$ and $p_{d_n}(t)$ are the power generation from the renewable energy sources and power demand in microgrid $n$, respectively.

For microgrid $n$ with energy storage system, then it satisfies

\vspace{-1mm}
\begin{equation}
\begin{array}{rcl}
p_{d_n}(t)=p_{g_n}(t)+p_{r_n}(t)+s_n(t)-s_n(t-1).
\end{array}
\end{equation}

\vspace{-0.6mm}
\begin{subequations}
\begin{equation*}  subject \ to \quad \forall\ t \in \mathcal{T} \end{equation*}
\begin{equation}  0 \leq s_n(t) \leq \overline{s}_n \end{equation}
\begin{equation}  \textbar s_n(t)-s_n(t-1) \textbar \leq \Delta s_n \end{equation}
\end{subequations}

\noindent where (3a) is the maximum storage capacity constraints, $s_n(t)$ is the stored energy of microgrid $n$ at time $t$. And $\overline{s}_n$ denotes the maximum capacity of the storage. (3b) is the maximum charging/discharging rate constraints, $\Delta s_n$ is the maximum charging/discharging power from time $t$ to $t+1$.

To consider the shiftable loads, the power demand $p_{d_n}(t)$ can be rewritten as 

\vspace{-5mm}
\begin{equation}
\begin{split}
p_{d_n}(t) &=f_{d_n}(\lambda(t),l_{b_n}(t)) = (1+h_n(\lambda(t)))l_{b_n}(t) \\
\end{split}
\end{equation}

\noindent where $\lambda(t)$ is the electricity price and $l_{b_n}(t)$ is the nominal value of the baseload at time $t$. Since the baseload has almost no fluctuations in practice and the baseload prediction technology can achieve high-precision prediction results, we will assume that $l_{b_n}(t)$ is a known data in advance.

Different household users may have different responses to the same price. The responses of different household users to various price schemes can be modelled by adopting a utility function from microeconomics \cite{16}. For each user, the utility function denotes the level of user's satisfaction corresponding to the energy consumption. The overall function of multi-microgrid can be expressed as

%Recent investigations indicate that certain utility functions can accurately model the behaviour of power users \cite{17}. The overall function of multi-microgrid can be expressed as
\vspace{-5mm}
\begin{equation}
\begin{split}
\max_{\lambda(t)}: F_w & =f_w(p_{d_1}(t),...,p_{d_N}(t),\lambda(t)) \\
& =\sum^{N}_{n=1}(f_u(p_{d_n}(t),\omega_n)-f_c(\lambda(t),p_{d_n}(t)))\end{split}
\end{equation} 

\noindent where
\vspace{-2mm}
\begin{subequations}
\begin{equation}  \begin{split}
f_c(\lambda(t),p_{d_n}(t))=\lambda(t)*p_{d_n}(t) 
\end{split} \end{equation}
\begin{equation}  \begin{split}
\ \ \ \ \ & \ \ \ \ \ \ \ \ \ \ f_u(p_{d_n}(t), \omega_n) \\
& =\left\{
\begin{aligned}
\omega_n p_{d_n}(t)-\frac{\alpha}{2}p_{d_n}(t)^2,\ \ &  if & \ 0\leq p_{d_n}(t) \leq \frac{\omega_n}{\alpha} \\
\frac{\omega_n}{\alpha}, \ \ \ \ \ \ \ \ \ \ \ & if & \ p_{d_n}(t) \geq \frac{\omega_n}{\alpha}
\end{aligned}
\right.
\end{split} \end{equation}
\end{subequations}

where $F_m$ is the overall welfare function. $f_c(\lambda(t),p_{d_n}(t))$ and $f_c(p_{d_n}(t), \omega_n)$ are the cost function imposed by the energy provider and the quadratic utility function of the user corresponding to linear decreasing marginal benefit, respectively. $p_{d_n}(t)=(1+h(\lambda(t))b_n(t))$ is the consuming power, $n=1,2,…,N$ is the number of microgrid. $b_n (t)$ is the base load and $h(\lambda(t)$ is the positive/negative value in percentage at different profiles of price signal. $\omega_n$ is the value which may change among users and also at different intervals of the day. $\alpha$ is a pre-determined parameter. For each advertised price value $\lambda(t)$, each user tries to modify the energy consumption to maximise its own welfare. It can be accomplished by setting the derivative of $F_m$ equal zero, which means that the marginal benefit of the user would be equal to the advertised price.

\subsection{ISO Model}

The ISOs in this paper mainly act as an emergency energy provider to afford emergency demand response programs. It will store as much energy as possible to achieve a secure level. To provide the maximum emergency energy and increase the life of the batteries, the utility function can be given as follows:

\vspace{-5mm}
\begin{equation}
\begin{split}
\max_{\lambda(t), p_{g_n}(t)}: F_s= \sum_{n=1}^{N_s}s_n(t).
\end{split}
\end{equation} 

\vspace{-2mm}
\begin{equation}
\begin{split}
 subject \ to \quad \forall\ t \in \mathcal{T}  \\
\underline{s}_n \leq s_n(t) \leq \overline{s}_n.
\end{split}
\end{equation}

\noindent where $\underline{s}_n$ is the minimum stored energy level for emergency operation, $n \in N_s$.

\subsection{Power Grid Model}

The power grid mainly injects the power to the microgrid when renewable energy in a microgrid is not enough. However, it can also absorb the power from the microgrid when renewable energy in microgrid has surplus energy. The mathematical models of the power grid can be written as

\vspace{-5mm}
\begin{equation}
\begin{split}
p_g(t)=\sum_{n=1}^{N}p_{g_n}(t).
\end{split}
\end{equation}

\noindent where $p_g(t)$ is the total power distribution from power grid and all microgrids. 

To obtain the maximum interest of the power grid, the derived from providing power $p_g(t)$ at the main power grid can be written as

\vspace{-5mm}
\begin{equation}
\begin{split}
\max_{\lambda(t), p_{g_n}(t)}: F_g & = f_g(\lambda(t), p_{g}(t))\\
& = \lambda(t)p_{g}(t)-C_{ p_g }(t)
\end{split}
\end{equation} 

\noindent where

\vspace{-5mm}
\begin{equation}
\begin{split}  
C_{ p_g }(t)=a_g p_g (t)^2+b_g p_g (t)+c_g
\end{split}
\end{equation} 

\noindent where $C_{ p_g }(t)$ is the quadratic cost functions. $a_g > 0$ and $b_g, c_g \geq 0$  are the pre-determined generator parameters.

The main research scenario, including demand and generation data based on the Penryn Campus, University of Exeter, was presented. The university office of general affairs will act as the ISO to purchase electrical energy from the power company and combine the existing RESs and energy storages to generate a new time-varying electricity price. Students living in the university student apartments need to pay their own electricity bills for the use of various electrical appliances, e.g., washing machine, dryers and freezer. If the electricity price changes at different times, students may change their electricity consumption behaviour to reduce their electricity bills. At the same time, the university can use time-varying electricity prices to reduce the load in a high electric period and optimise the operation of energy storage systems to reduce the purchase of electricity from the main grid. A number of smart meters installed are 4.7 million and 4.5 million in British homes in 2018 and 2019, respectively \cite{21}. Therefore, this designed scenario is also very practical in a smart grid environment in the local community.

\section{Multi-Objective Problem Formulation}

This section will describe a multi-objective problem formulation to maximise the benefits of three participants for multi-microgrid system design. A multi-objective reinforcement learning (MORL) technique is then proposed to optimise the problem in a real-time market scenario.

To solve these three objectives $F_w$, $F_s$ and $F_g$ at the same time, a MOP formulation is given as
\vspace{-2mm}
\begin{subequations}
\begin{equation}  \min_{\lambda(t)} \  -F_w=-f_w(p_{d_1}(t),...,p_{d_N}(t),\lambda(t)) \end{equation}
\begin{equation}  \min_{\lambda(t), p_{g_n}(t)} \ -F_s=-\sum_{n=1}^{N_s}s_n(t) \end{equation}
\begin{equation}  \min_{\lambda(t), p_{g_n}(t)} \ -F_g=-f_g(\lambda(t), p_{g}(t)) \end{equation}
\end{subequations}

\begin{subequations}
\begin{equation*}  subject \ to \quad (1)-(4),(8) \ and \ (9) \end{equation*}
\end{subequations}

\noindent where $\lambda(t)$ and $p_{g_n}(t)$ are the two variables related to the ISO determined by the current generation of renewable energy and the status of energy storage during the period. To solve the problem considering all the constraints, an additional function is introduced in the following way

\vspace{-2mm}
\begin{equation}
\begin{split}
F_a  = &\sum_{n=1}^{N_s}[max(|s_n(t)-s_n(t-1)-\Delta s_n,0) \\
    & +  max(s_n(t)-\overline{s}_n,0) + max(\underline{s}_n-s_n(t),0)]
\end{split}
\end{equation} 

\noindent where $F_a$ is determined by the stored energy. The multi-microgrid is stable and satisfies the all constraints if and only if $F_a=0$. According to the equation (14), the resulting MOP (12) can be rewritten by

\vspace{-5mm}
\begin{equation}
\begin{split}
\min_{\lambda(t), p_{g_n}(t)} \ F_m=[-F_w \ -F_s \ -F_g \ \ F_a]^T
\end{split}
\end{equation} 

To solve the MOP, the Pareto optimality will be utilised to satisfy all requirements. In particular, evolutionary algorithms have been widely recognized as a major approach for MOPs~\cite{LiZZL09,LiZLZL09,CaoWKL11,LiKWCR12,LiKCLZS12,LiKWTM13,LiK14,CaoKWL14,WuKZLWL15,LiKZD15,LiKD15,LiDZ15,LiDZZ17,WuKJLZ17,WuLKZZ17,LiDY18,ChenLY18,ChenLBY18,LiCFY19,WuLKZZ19,LiCSY19,Li19,GaoNL19,LiXT19,ZouJYZZL19,LiLDMY20,WuLKZ20,BillingsleyLMMG19,LiX0WT20}.
%
%The detailed discussion can be found as follows.
%
%\textit{Definition - Pareto Dominance:} Let $\textbf{H(x)}$ be a vector function of MOP and a feasible solution space is $\Omega$. The MOP is to find a vector $u \in \Omega$ that optimises the vector function. A vector $u$ dominates $u^{'}$ (denoted by $u \prec u^{'}$) if the condition $H_i(u) \leq H_i(u^{'})$ holds true for all $i$ and there is at least one $i$ satisfy $H_i(u) < H_i(u^{'})$. It means that one solution is better than other if it is better in one objective and in the other is equal.
%
%\textit{Definition - Pareto Optimal:} A vector $u^*$ is Pareto optimal if there does not exist a feasible solution $u^{'} \prec u^*$ in the solution space that dominates it. 
%
%\textit{Definition - Pareto Optimal Set:} The Pareto optimal set for a MOP is defined as the set of Pareto optimal solutions, where $P^*=\{u^* \in \Omega\}$.
%
%\textit{Definition - Pareto Front:}  The boundary defined by the set of all point mapped from the Pareto optimal set is called the Pareto front.

\section{Proposed Algorithm for Multi-Microgrid Optimisation}

To obtain the Pareto optimal set for MOP, a multi-objective Q learning algorithm is proposed in this section. The MORL framework is based on a scalarized single-policy algorithm that employs scalarization functions to reduce the dimensionality of the multi-objective environment to a single and scalar dimension.

A scalarization function can be defined as
\vspace{-2mm}
\begin{equation}
\begin{split}
F=f(\textbf{x},\textbf{w})
\end{split}
\end{equation} 

\noindent where in the case of MORL, $\textbf{x}$  and $\textbf{w}$ in the objective functions are the \textbf{Q} vector and the weight vector, respectively. The scalar Q values can be extended to \textbf{Q} vector that contains each Q value for each objective. When an action is chosen, the function $F$ will be applied to the \textbf{Q} vector to achieve a single, scalar $SQ(s,a)$ estimate.
%
%\begin{equation}
%\begin{split}
%\textbf{Q}(s,a)=(Q_1(s,a),...,Q_m(s,a))
%\end{split}
%\end{equation} 
%
%\noindent where $s$ and $a$ are the state and action, respectively. When an action is chosen, the function $v_{\textbf{w}}$ will be applied to the \textbf{Q} vector to achieve a single, scalar $SQ(s,a)$ estimate.
\vspace{-2mm}
\begin{equation}
\begin{split}
SQ(s,a)=\sum_{n=1}^{N}{w}_n \cdot {Q}_n(s,a)
\end{split}
\end{equation} 

\noindent where $n \in N$ is stand for the index of each objective function. And the weight vector should satisfy the following equation $ \sum_{n=1}^{N}{w}_n=1 $.

However, this linear scalarization function has a fundamental limitation that can only find policies in convex regions of the Pareto optimal set \cite{18}. Then a scalarization function based on the $L_p$ metrics is proposed in this paper \cite{19}. The proposed $L_p$ metrics measure the distance between a point $\textbf{x}$ in the multi-objective space and a utopian point $\textbf{z}^{*} $ which is an adjustable parameter during the learning process. The measured distance between $\textbf{x} $ and $ \textbf{z}^* $ for each objective can be given as follows

\vspace{-5mm}
\begin{equation}
\begin{split}
L_p(x)=(\sum_{n=1}^{N}{w}_n |x_n-z_n^*|^p)^{1/p}.
\end{split}
\end{equation} 

\noindent where $1 \leq p \leq \infty $. In the case of $p = \infty$, the metric can be known as the weighted $L_{\infty}$ or the Chebyshev metric

\vspace{-5mm}
\begin{equation}
\begin{split}
L_{\infty}(x)=\max_{n=1,...,N}w_n |x_n-z_n^*|.
\end{split}
\end{equation} 

\noindent where in the case of MORL, $x_n$ can be replaced by $Q_n(s,a)$ to obtain the $SQ(s,a)$ with state $s$ and action $a$

\vspace{-5mm}
\begin{equation}
\begin{split}
SQ(s,a)=\max_{n=1,...,N}w_n |Q_n(s,a)-z_n^*|.
\end{split}
\end{equation} 

RL elements, including state and action spaces, reward function, learning and exploration rates, and discount factor, are described in detail in the following subsections:

\vspace{-3mm}
\subsubsection{\textbf{State Space}} The state variables are time of day ($ToD_j$) and State of Charge ($SoC_k$). 

\vspace{-7mm}
\begin{equation}
\begin{split}
s|s_{j,k}=(ToD_j, SoC_k)\\
\end{split}
\end{equation} 

\noindent where the time of day $ToD$ is discretised into 24 hours $j=1,2,...,24$, the State of Charge $SoC$ is divided into 8 levels from 30$\%$ to 100$\%$.

\vspace{-4mm}
\subsubsection{\textbf{Action Space}} Action Space is the combination of Price and Charging/Discharging.

\vspace{-1mm}
\begin{equation}
\begin{array}{rcl}
A=\{a|(Price Signal, charging/discharging/idle)\}
\end{array}
\end{equation}

\noindent where a total of 24 actions can be selected. The price is divided into 8 values which are set from 1.5 to 5.0.

\vspace{2mm}
\begin{tabular}{l}

\hline
\textbf{Algorithm 1}: Scalarized $\epsilon$ greedy strategy\\

\hline
1: \ \textbf{Initialise} \textit{SQList}\\
2: \ \textbf{for} each action $a \in A$ \textbf{do}\\ 
3: \quad \quad \textbf{x} $\leftarrow$  $\{Q_1(s,a),...,Q_m(s,a)\}$ \\
4: \quad \quad \textit{SQ(s,a)} $\leftarrow$  $f(\textbf{x},\textbf{w})$ \\
5: \quad \quad Append \textit{SQ(s,a)} to  \textit{SQList} \\
6: \ \textbf{End for}\\
7: \ \textbf{return $\epsilon$ greedy(\textit{SQList})}\\
\hline
\end{tabular}

%\vspace{-5mm}

\vspace{1mm}
\begin{tabular}{l}

\hline
\textbf{Algorithm 2}: Multi-objective Q-learning algorithm)\\

\hline
1: \ \textbf{Initialise} $Q_n(s,a)$ \\
2: \ \textbf{for} each episode \textbf{do} \\ 
3: \quad \quad Initialise state $s$\\
4: \quad \quad \textbf{repeat} \\
5: \quad \quad \quad Select action $a$ using $\epsilon$ greedy strategy \\ 
6: \quad \quad \quad Take action and observe new state $s^{'} \in S$ \\
7: \quad \quad \quad Obtain reward vector $\textbf{r}$ and select new action \\
8: \quad \quad \quad \textbf{for} each objective $n$ \textbf{do} \\
9: \quad \quad \quad \quad $Q_n(s,a)=Q_n(s,a)+\alpha_t(r_n+\gamma Q_n(s^{'},a^{'})$ \\
\ \ \quad \quad \quad \quad \quad \quad \quad \quad \quad \  $-Q_n(s,a))$ \\ 
10: \, \, \, \quad \textbf{end for}\\
11: \, \, \, \quad $s \rightarrow s^{'}$ \\
12: \, \, \, \textbf{until} $s$ is terminal\\
13: \textbf{end for}\\
\hline
\end{tabular}

\vspace{-2mm}

\subsubsection{\textbf{Reward}} The reward value $r_n(t)$ for each objective is the immediate incentive gained by taking a specific action at state $s$. The reward function of each objective is designed to minimise the objective function. Then these obtained reward values will be saved to the extended $Q$ table.

According to these $SQ(s,a)$ values, an appropriate action selection strategy, i.e., scalarised $\epsilon$ greedy strategy, can be taken to select an action. The detailed scalarised $\epsilon$ greedy strategy can be found in Algorithm 1.

The proposed multi-objective Q-learning algorithm is shown in Algorithm 2. First of all, the Q values for each objective are initialised. Then the algorithm starts each episode in state $s$ and select action via Scalarized $\epsilon$ greedy strategy. In terms of the selected action, the algorithm will transit to a new state $s^{'}$ and produce the reward vector $\textbf{r}$. More precisely, these reward values are updated for each objective individually and single objective reward update rule will be extended to a multi-objective environment. Then the best scalarised action for the next state $s^{'}$ will be chosen via $\epsilon$ greedy.
As long as each action and state is fully sampled, convergence can be guaranteed.

\section{Simulation Results and Performance}

The numerical simulation results will be performed to assess the performance of the proposed multi-objective reinforcement learning algorithm. It is assumed there are three microgrids $\mathcal{N}=3$ and two microgrids with energy storage $\mathcal{N_s}=2$. The two maximum storage capacities are 200 kWh and 250 kWh, respectively. Let $\Delta s_n$ equals 10$\%$ of the maximum capacity. The typical power demand corresponding to price $\lambda$ can be obtained as discussed in \cite{20}. And the baseload $l_{b_n}$ from the Penryn Campus, Exeter University. An example of baseload and generation on Nov 17, 2019 is shown in Fig. 1. The power demand can increase/decrease in terms of the price value when the price signal varied.

\vspace{-8mm}
\begin{figure}[htbp]

\centering

\includegraphics [width=3.5in]{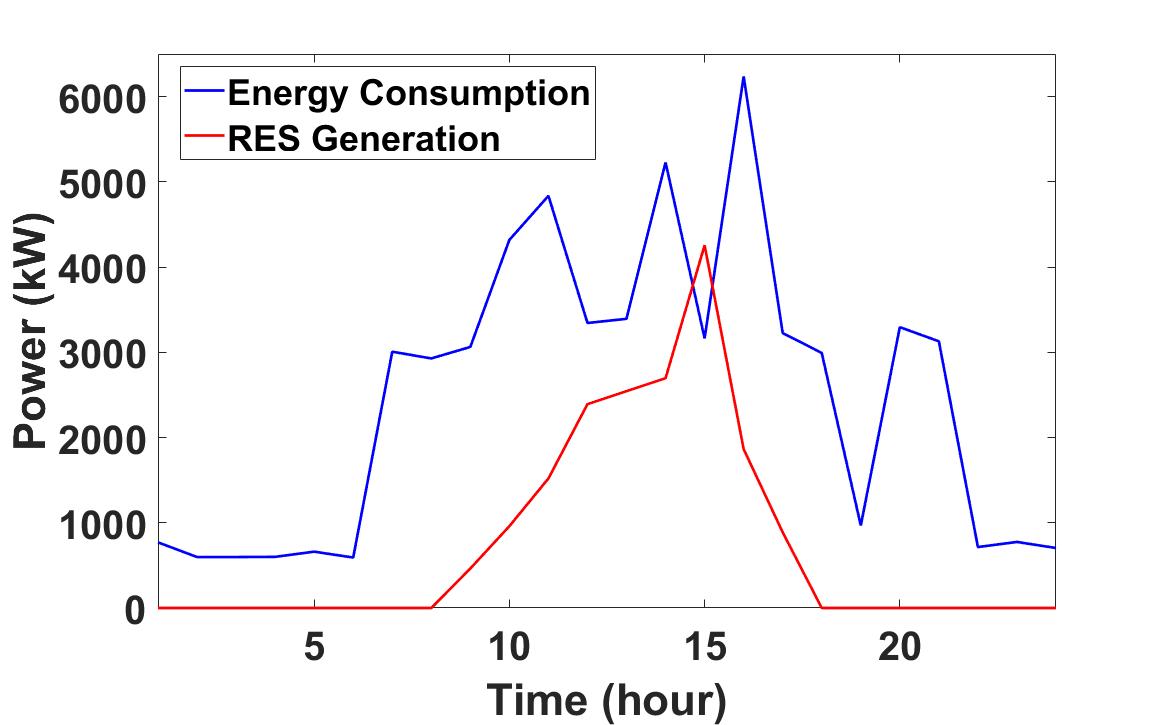}

\caption{Demand and RES generation of Microgrid 1 on Nov 17, 2019. } \label{fig:5}

\end{figure}

\vspace{-10mm}
\begin{figure}[htbp]

\centering

\includegraphics [width=3.5in]{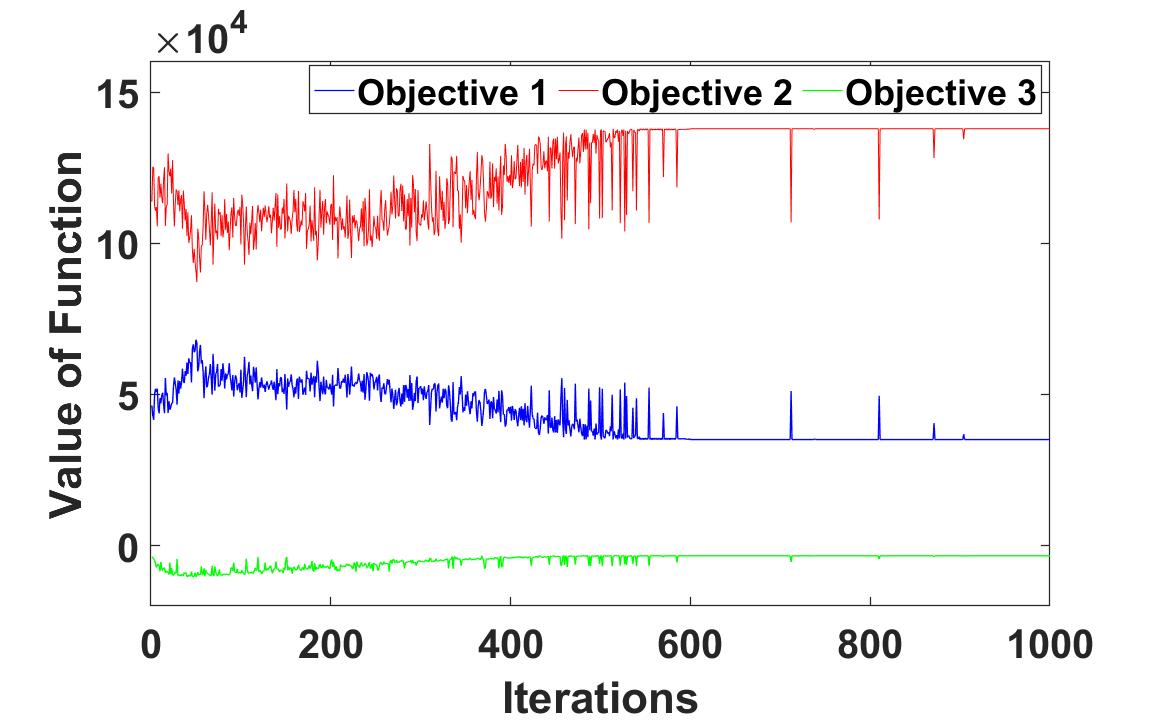}

\caption{Sample of Convergence Rate of MORL. } \label{fig:5}

\end{figure}

%\vspace{-5mm}
%
%\vspace{-5mm}
Fig. 2 show an example of an extreme value of Approximated Pareto Front (APF) that minimises the first objective function $-F_w$. However, the other two functions are affected by objective function one and cannot converge to the minimum. Obviously, each objective function will affect each other, so it is necessary to provide a fair design for all participants.

One particular solution $P^*$ in Pareto optimal set that can maximise the minimum improvement in all dimensions is discussed. Fig. 3 shows an example of the APF and reveals the connection between these three objectives. Three distinct solutions $p^*_1$, $p^*_2$ and $p^*_3$ are the extreme dominated solutions for three objective functions, respectively. It means that each solution will advantage to every single objective function. To ensure fairness for all objective functions, one particular solution $P^*$ based on the APF will be picked to give no advantage to any single objective. As can be seen in Fig. 3, the optimal solution is located in the centre of the APF graphically. It refers that the proposed MORL approach can provide a fair solution to all three participants.

\vspace{-1em}
\begin{figure}[htbp]

\centering

\includegraphics [width=3.5in]{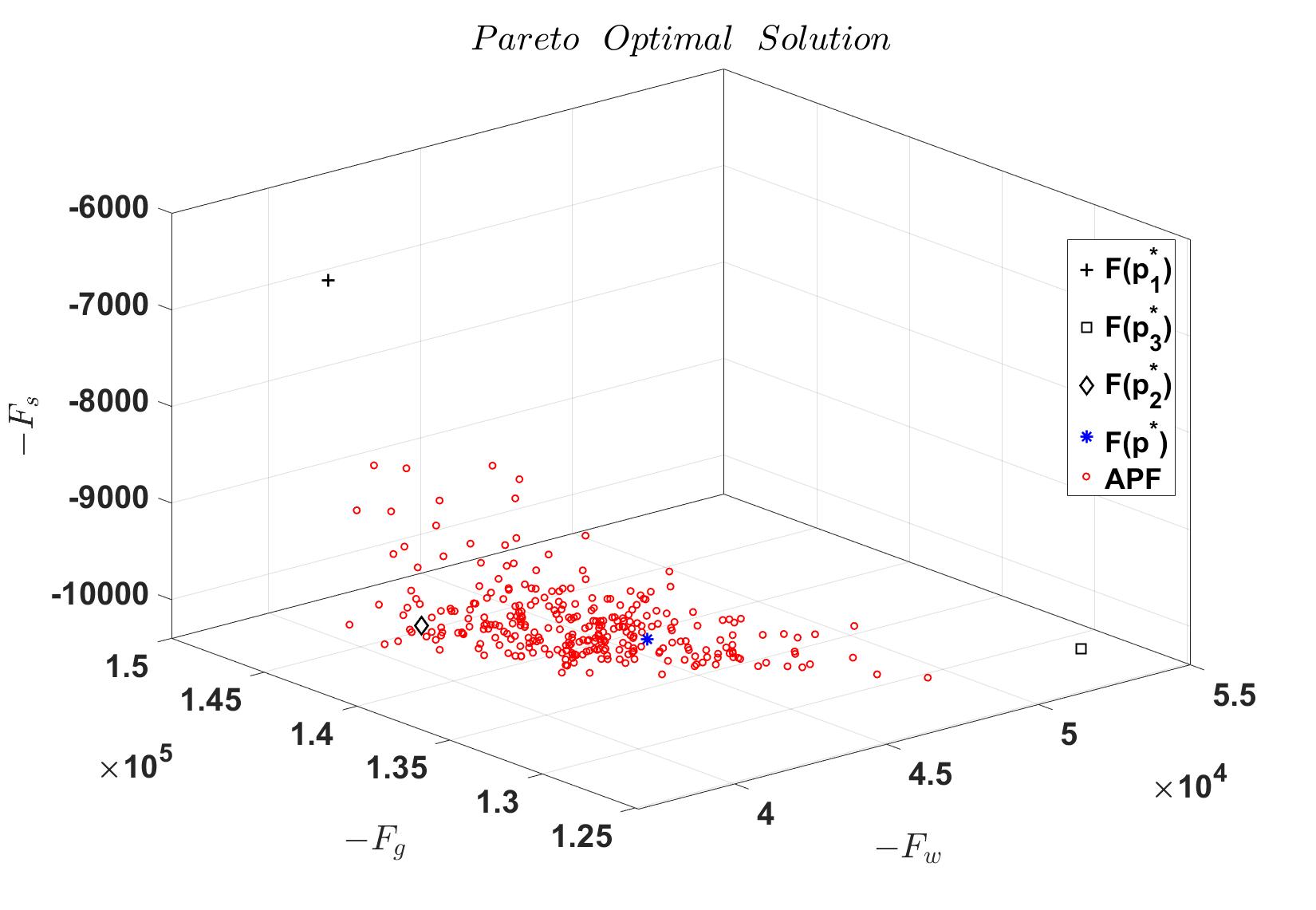}

\caption{APFs and non-dominated vectors $F(p^*)$ sampled. } \label{fig:5}

\end{figure}
\vspace{-5mm}

The fluctuation of price signal plays a significant role in smart grid energy management. Fig. 4 proves that the proposed method can generate outstanding price fluctuation. Ideally, high prices will produce peak load reduction and discharge energy storage, while low prices will fill valley load and charge energy storage. In this multi-objective scenario, since all three participants want to maximise their own benefits, for example, emergency energy provider does not mind the price of electricity and only consider the secure energy levels. Thus, only a small portion of the demand may be shifted for some Pareto optimal solutions. 

\vspace{-1em}
\begin{figure}[htbp]

\centering

\includegraphics [width=3.5in]{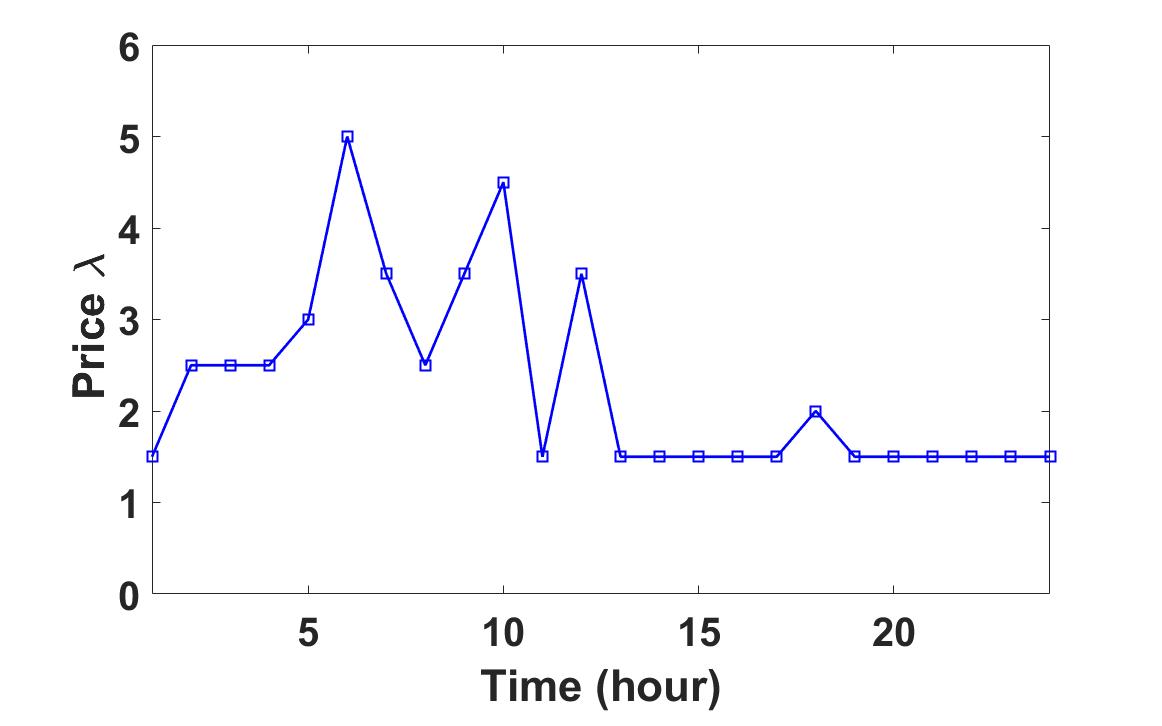}

\caption{Dynamic price signal $\lambda$ by using the proposed MORL approach. } \label{fig:5}

\end{figure}
\vspace{-1em}

The Pareto optimal set from the experimental results in Fig. 3 shows that the proposed MORL is able to benefit to a singular participant or balance to all participants.  Since the learned Q values encompass the agents' past experiences without re-solving the decision problem, it can solve the multi-objective problems faster than traditional optimisation algorithms.  In summary, all the experimental results verify the performance of the MORL. It is capable of managing the multi-objective smart grid system design effectively and efficiently.

\vspace{-1em}
\section{Conclusion}
\vspace{-1em}

This paper proposes a multi-microgrid planning model that considers the dynamic electricity prices and renewable energy sources. The planning scenario is analysed by the multi-objective reinforcement learning algorithm, which optimises the electricity price and the operation of energy storage. Meanwhile, the dynamic prices of the multi-microgrid system are determined by the power demand from the main grid, which considers the benefit of all three participants. The simulation results show that the proposed MORL algorithm can provide a fair and effectiveness operating planning for all participants by control the energy storage operation and modify the real-time energy tariff. It shows the ability of MORL to learn the optimal control policy. The optimised coordinated operation can help to improve the utilisation of renewable energy, increase the operating life of batteries, reduce the operation cost of multi-microgrid, save bills for customers and maximise the profit for the power grid.

\section*{Acknowledgement}
K. Li was supported by UKRI Future Leaders Fellowship (Grant No. MR/S017062/1).

\bibliographystyle{IEEEtran}
\bibliography{IEEEabrv,reference11}

\end{document}